\documentclass{article}
\usepackage{ijcai17}
\usepackage{epstopdf}
\usepackage{times}
\usepackage{graphicx}
\usepackage{subfigure}
\usepackage{multirow}
\usepackage{enumitem}
{\tiny {\footnotesize {\tiny }}}
\usepackage[labelfont=bf, textfont=bf]{caption}
\usepackage{graphicx}
\usepackage{amsfonts,amssymb}
\usepackage{amsmath}
\usepackage{caption}
\DeclareCaptionFont{9pt}{\fontsize{9pt}{10pt}\selectfont}
\usepackage{url}
\makeatletter
\def\url@leostyle{%
	\@ifundefined{selectfont}{\def\UrlFont{\sf}}{\def\UrlFont{}}}
\makeatother
\newcommand{\ie}{\emph{i.e., }}
\newcommand{\eg}{\emph{e.g., }}

\newcommand{\wrt}{\emph{w.r.t. }}
\newcommand{\cf}{\emph{cf. }}
\newcommand{\aka}{\emph{aka. }}

\title{Attentional Factorization Machines: \\Learning the Weight of Feature Interactions via Attention Networks\thanks{The corresponding author is Xiangnan He.}}

\author{
	Jun Xiao$^{1}$\qquad 
	Hao Ye$^{1}$\qquad 
	Xiangnan He$^{2}$\qquad
	Hanwang Zhang$^{2}$\qquad 
	Fei Wu$^{1}$\qquad
	Tat-Seng Chua$^{2}$\\
	$^{1}$College of Computer Science, Zhejiang University\\
	$^{2}$School of Computing, National University of Singapore\\
	{\{junx, wufei\}@cs.zju.edu.cn \quad \{xiangnanhe, haoyev, hanwangzhang\}@gmail.com \quad dcscts@nus.edu.sg}
}

\begin{document}

\maketitle

\begin{abstract}
	\textit{Factorization Machines} (FMs) are a supervised learning approach that enhances the linear regression model by incorporating the second-order feature interactions. Despite effectiveness, FM can be hindered by its modelling of all feature interactions with the same weight, as not all feature interactions are equally useful and predictive.  
	For example, the interactions with useless features may even introduce noises and adversely degrade the performance. In this work, we improve FM by discriminating the importance of different feature interactions. We propose a novel model named \textit{Attentional Factorization Machine} (AFM), which learns the importance of each feature interaction from data via a neural attention network. Extensive experiments on two real-world datasets demonstrate the effectiveness of AFM.
	Empirically, it is shown on regression task AFM betters FM with a $8.6\%$ relative improvement, and consistently outperforms the state-of-the-art deep learning methods Wide\&Deep~\cite{cheng2016wide} and DeepCross~\cite{shan2016deep} with a much simpler structure and fewer model parameters. Our implementation of AFM is publicly available at: \url{https://github.com/hexiangnan/attentional_factorization_machine}
\end{abstract}

\section{Introduction}

Supervised learning is one of the fundamental tasks in machine learning (ML) and data mining. The goal is to infer a function that predicts the target given predictor variables (\aka features) as input. For example, real valued targets for regression and categorical labels for classification.
It has broad applications including recommendation systems~\cite{iCD,zhao2016user}, online advertising~\cite{shan2016deep,juan2016field}, and image recognition~\cite{zhang2017relation,wang2015visual}. 

When performing supervised learning on categorical predictor variables, it is important to account for the interactions between them~\cite{He_NFM,cheng2016wide}. 
As an example, let us consider the toy problem of predicting customers' income with three categorical variables: 1) $occupation$ = \{banker,engineer,...\}, 2) $level$ = \{junior,senior\}, and 3) $gender$ = \{male,female\}. 
While junior bankers have a lower income than junior engineers, it can be the other way around for customers of senior level --- senior bankers generally have a higher income than senior engineers. 
If a ML model assumes independence between predictor variables and ignores the interactions between them, it will fail to predict accurately, such as linear regression that associates a weight for each feature and predicts the target as the weighted sum of all features. 

To leverage the interactions between features, one common solution is to explicitly augment a feature vector with products of features (\aka cross features), as in polynomial regression (PR) where a weight for each cross feature is also learned.
However, the key problem with PR (and other similar cross feature-based solutions, such as the wide component of Wide\&Deep~\cite{cheng2016wide}) is that for sparse datasets where only a few cross features are observed, the parameters for unobserved cross features cannot be estimated. 

To address the generalization issue of PR, factorization machines (FMs)\footnote{In this paper, we focus on the second-order FM, which is the most effective and widely used instance of FMs.} were proposed~\cite{FM}, which parameterize
the weight of a cross feature as the inner product of the embedding vectors of the constituent features. 
By learning an embedding vector for each feature, FM can estimate the weight for any cross feature. 
Owing to such generality, FM has been successfully applied to various applications, ranging from recommendation systems~\cite{silkroad,Chen:2016} to natural language processing~\cite{petroni2015core}. 
Despite great promise, we argue that FM can be hindered by its modelling of all factorized interactions with the same weight.
In real-world applications, different predictor variables usually have different predictive power, and 
not all features contain useful signal for estimating the target,
such as the $gender$ variable for predicting customers' income in the previous example. 
As such, the interactions with less useful features should be assigned a lower weight as they contribute less to the prediction. 
Nevertheless, FM lacks such capability of differentiating the importance of feature interactions, which may result in suboptimal prediction.

In this work, we improve FM by discriminating the importance of feature interactions. 
We devise a novel model named AFM, which utilizes the recent advance in neural network modelling --- the attention mechanism~\cite{Attentive_CF,chen2017cvpr} --- to enable feature interactions contribute differently to the prediction. 
More importantly, the importance of a feature interaction is automatically learned from data without any human domain knowledge. 
We conduct experiments on two public benchmark datasets of context-aware prediction and personalized tag recommendation. 
Extensive experiments show that our use of attention on FM serves two benefits: it not only leads to better performance, but also provides insight into which feature interactions contribute more to the prediction.
This greatly enhances the interpretability and transparency of FM, allowing practitioners to perform deeper analysis of its behavior. 



\section{Factorization Machines}
\label{sec:FM}
As a general ML model for supervised learning, factorization machines were originally proposed for collaborative recommendation~\cite{FM,fastFM}. Given a real valued feature vector $\textbf{x}\in \mathbb{R}^n$ where $n$ denotes the number of features, FM estimates the target by modelling all interactions between each pair of features:
\begin{equation}
	\hat{y}_{FM}(\textbf{x}) = \underbrace{w_0 + \sum_{i=1}^n w_i x_i}_{\text{linear regression}} + \underbrace{\sum_{i=1}^n\sum_{j=i+1}^n \hat{w}_{ij}x_i x_j}_{\text{pair-wise feature interactions}},
\end{equation}
where $w_0$ is the global bias, $w_i$ denotes the weight of the $i$-th feature, and $\hat{w}_{ij}$ denotes the weight of the cross feature $x_i x_j$, which is factorized as:
$
	\hat{w}_{ij} = \textbf{v}_i^T \textbf{v}_j,
$
where $\textbf{v}_i\in \mathbb{R}^k$ denotes the embedding vector for feature $i$, and $k$ denotes the size of embedding vector. Note that due to the coefficient $x_i x_j$, only interactions between non-zero features are considered. 

It is worth noting that FM models all feature interactions in the same way: first, a latent vector $\textbf{v}_i$ is shared in estimating all feature interactions that the $i$-th feature involves; second, all estimated feature interactions $\hat{w}_{ij}$ have a uniform weight of $1$. In practice, it is common that not all features are relevant to prediction. As an example, consider the problem of news classification with the sentence ``US continues taking a leading role on foreign payment transparency''. 
It is obvious that the words besides ``foreign payment transparency'' are not indicative of the topic of the (financial) news. 
Those interactions involving irrelevant features can be considered as noises that have no contribution to the prediction. 
However, FM models all possible feature interactions with the same weight, which may adversely deteriorate its generalization performance. 

\section{Attentional Factorization Machines}
\label{sec:AFM}

\begin{figure*}[t]
	\centering
	\includegraphics[scale=0.45]{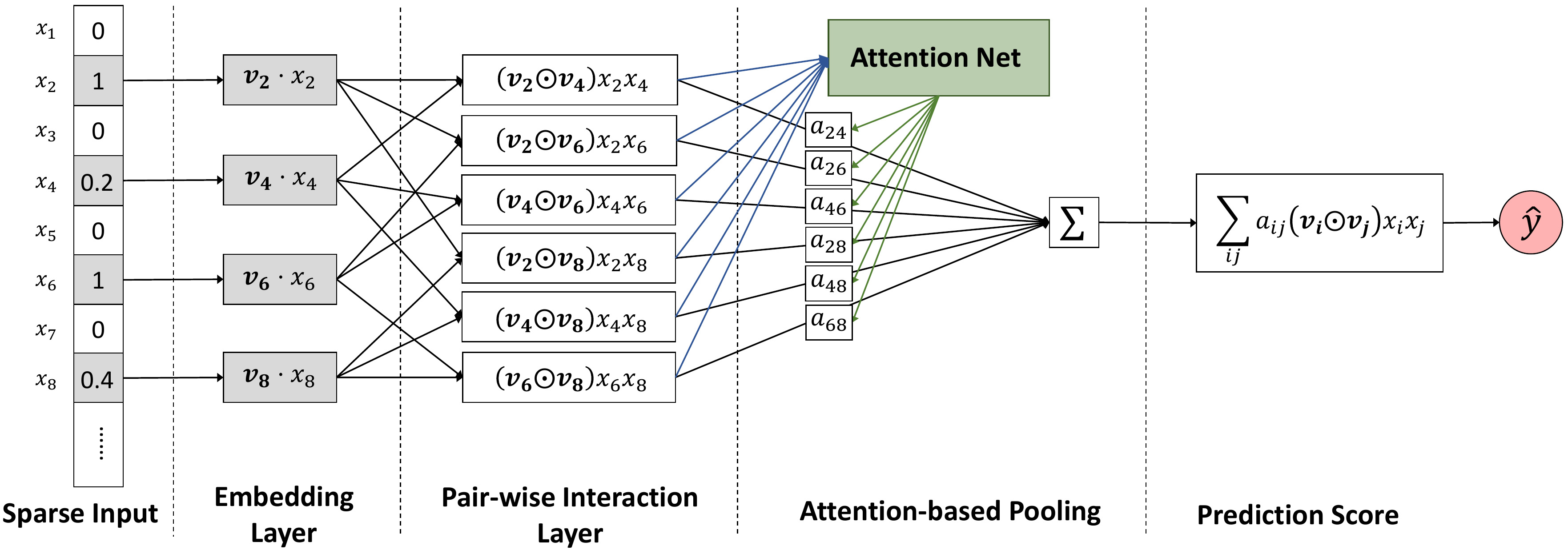}
	\vspace{-5pt}
	\caption{The neural network architecture of our proposed Attentional Factorization Machine model.}
	\vspace{-10pt}
	\label{fig:nfm}
\end{figure*}


\subsection{Model}
Figure~\ref{fig:nfm} illustrates the neural network architecture of our proposed AFM model.
For clarity purpose, we omit the linear regression part in the figure, which can be trivially incorporated.  
The input layer and embedding layer are the same with FM, which adopts a sparse representation for input features and embeds each non-zero feature into a dense vector. In the following, we detail the pair-wise interaction layer and the attention-based pooling layer, which are the main contribution of this paper. 

\subsubsection{Pair-wise Interaction Layer}
Inspired by FM that uses inner product to model the interaction between each pair of features, we propose a new \textit{Pair-wise Interaction Layer} in neural network modelling. It expands $m$ vectors to $m(m-1)/2$ interacted vectors, where each interacted vector is the element-wise product of two distinct vectors to encode their interaction. 
Formally,  let the set of non-zero features in the feature vector $\textbf{x}$ be $\mathcal{X}$, and the output of the embedding layer be $\mathcal{E} = \{\textbf{v}_i x_i\}_{i\in \mathcal{X}}$. 
We can then represent the output of the pair-wise interaction layer as a set of vectors:
\begin{equation}
	f_{PI}(\mathcal{E}) = \{(\textbf{v}_i\odot \textbf{v}_j) x_i x_j \}_{(i,j)\in\mathcal{R}_x},
\end{equation}
where $\odot$ denotes the element-wise product of two vectors, and $\mathcal{R}_x = \{(i,j)\}_{i\in \mathcal{X}, j\in \mathcal{X}, j>i}$ for short. By defining the pair-wise interaction layer, we can express FM under the neural network architecture. 
To show this, we first compress $f_{PI}(\mathcal{E})$ with a sum pooling, and then use a fully connected layer to project it to the prediction score:
\begin{equation}
	\hat{y} = \textbf{p}^T \sum_{(i,j)\in\mathcal{R}_x} (\textbf{v}_i\odot \textbf{v}_j) x_i x_j + b,
\end{equation}
where $\textbf{p}\in\mathbb{R}^k$ and $b\in\mathbb{R}$ denote the weights and bias for the prediction layer, respectively. Clearly, by fixing $\textbf{p}$ to $\textbf{1}$ and $b$ to $0$, we can exactly recover the FM model. Note that our recent work of neural FM has proposed a \textit{Bilinear Interaction} pooling operation~\cite{He_NFM}, which can be seen as using a sum pooling over the pair-wise interaction layer. 

\subsubsection{Attention-based Pooling Layer}
Since the attention mechanism has been introduced to neural network modelling, it has been widely used in many tasks, such as recommendation~\cite{Attentive_CF}, information retrieval~\cite{xiong_attend}, and computer vision~\cite{chen2017cvpr}. 
The idea is to allow different parts contribute differently when compressing them to a single representation. Motivated by the drawback of FM, we propose to employ the attention mechanism on feature interactions by performing a weighted sum on the interacted vectors: 
\begin{equation}
	f_{Att}(f_{PI}(\mathcal{E})) = \sum_{(i,j)\in\mathcal{R}_x} a_{ij}(\textbf{v}_i\odot \textbf{v}_j) x_i x_j,
\end{equation}
where $a_{ij}$ is the attention score for feature interaction $\hat{w}_{ij}$, which can be interpreted as the importance of $\hat{w}_{ij}$ in predicting the target. To estimate $a_{ij}$, an intuitive solution to directly learn it by minimizing the prediction loss, which also seems to be technically viable. However, the problem is that, for features that have never co-occurred in the training data, the attention scores of their interactions cannot be estimated. To address the generalization problem, we further parameterize the attention score with a multi-layer perceptron (MLP), which we call the \textit{attention network}. 
The input to the attention network is the interacted vector of two features, which encodes their interaction information in the embedding space. Formally, the attention network is defined as:
\begin{equation}
\label{eq:a_ij}
\begin{aligned}
	a'_{ij} &= \textbf{h}^T ReLU(\textbf{W} (\textbf{v}_i \odot \textbf{v}_j)x_i x_j + \textbf{b}), \\
	a_{ij} &= \frac{\exp(a'_{ij})}{\sum_{(i,j)\in \mathcal{R}_x } \exp(a'_{ij})}, \\
\end{aligned}
\end{equation}
where $\textbf{W}\in \mathbb{R}^{t\times k}, \textbf{b}\in \mathbb{R}^{t}, \textbf{h}\in \mathbb{R}^{t}$ are model parameters, and $t$ denotes the hidden layer size of the attention network, which we call \textit{attention factor}. 
The attention scores are normalized through the softmax function, a common practice by previous work.
We use the rectifier as the activation function, which empirically shows good performance. 

The output of the attention-based pooling layer is a $k$ dimensional vector, which compresses all feature interactions in the embedding space by distinguishing their importance. We then project it to the prediction score. To summarize, we give the overall formulation of AFM model as:
\begin{equation}
	\hat{y}_{AFM}(\textbf{x}) = w_0 + \sum_{i=1}^n w_i x_i + \textbf{p}^T\sum_{i=1}^n\sum_{j=i+1}^n a_{ij}(\textbf{v}_i\odot\textbf{v}_j)x_i x_j,
\end{equation}
where $a_{ij}$ has been defined in Equation (\ref{eq:a_ij}). The model parameters are $\Theta = \{w_0, \{w_i\}_{i=1}^n, \{\textbf{v}_i\}_{i=1}^n, \textbf{p}, \textbf{W}, \textbf{b}, \textbf{h}\}$. 

\subsection{Learning}
As AFM directly enhances FM from the perspective of data modelling, it can also be applied to a variety of prediction tasks, including regression, classification and ranking. 
Different objective functions should be used to tailor the AFM model learning for different tasks. 
For regression task where the target  $y(\textbf{x})$ is a real value, a common objective function is the squared loss:
\begin{equation}
\label{eq:regression}
	L_r =\sum_{x \in \mathcal{T}} (\hat{y}_{AFM}(\textbf{x}) - y(\textbf{x}))^2,
\end{equation}
where $\mathcal{T}$ denotes the set of training instances. 
For binary classification or recommendation task with implicit feedback~\cite{He:WWW2017},
we can minimize the log loss.
In this paper, we focus on the regression task and optimize the squared loss. 

To optimize the objective function, we employ stochastic gradient descent (SGD) --- a universal solver for neural network models. 
The key to implement a SGD algorithm is to obtain the derivative of the prediction model $\hat{y}_{AFM}(\textbf{x})$ \wrt each parameter. As most modern toolkits for deep learning have provided the functionality of automatic differentiation, such as Theano and TensorFlow, we omit the details of derivatives here. 

\subsubsection{Overfitting Prevention} 
Overfitting is a perpetual issue in optimizing a ML model. It is shown that FM can suffer from overfitting~\cite{fastFM}, so the $L_2$ regularization is an essential ingredient to prevent overfitting for FM. As AFM has a stronger representation ability than FM, it may be even easier to overfit the training data. 
Here we consider two techniques to prevent overfitting --- dropout and $L_2$ regularization --- that have been widely used in neural network models. 

The idea of dropout is randomly drop some neurons (along their connections) during training~\cite{srivastava2014dropout}. It is shown to be capable of preventing complex co-adaptations of neurons on training data.
Since AFM models all pair-wise interactions between features while not all interactions are useful, the neurons of the pair-wise interaction layer may easily co-adapt with each other and result in overfitting. As such, we employ dropout on the pair-wise interaction layer to avoid co-adaptations. 
Moreover, as dropout is disabled during testing and the whole network is used for prediction, dropout has another side effect of performing model averaging with smaller neural networks, which may potentially improve the performance~\cite{srivastava2014dropout}. 

For the attention network component which is a one-layer MLP, we apply $L_2$ regularization on the weight matrix $\textbf{W}$ to prevent the possible overfitting. That is, the actual objective function we optimize is:
\begin{equation}
	L = \sum_{x \in \mathcal{T}} (\hat{y}_{AFM}(\textbf{x}) - y(\textbf{x}))^2 + \lambda ||\textbf{W}||^2,
\end{equation}
where $\lambda$ controls the regularization strength. We do not employ dropout on the attention network, as we find the joint use of dropout on both the interaction layer and attention network leads to some stability issue and degrades the performance. 
\section{Related Work}
\label{sec:related}
FMs~\cite{FM} are mainly used for supervised learning under sparse settings; for example, in situations where categorical variables are converted to sparse feature vector via one-hot encoding. 
Distinct from the continuous raw features found in images and audios, input features of the Web domain are mostly discrete and categorical~\cite{He_NFM}. 
For prediction with such sparse data, it is crucial to model the interactions between features~\cite{shan2016deep}.
In contrast to matrix factorization (MF) that models the interaction between two entities only~\cite{fastMF}, FM is designed to be a general machine learner for modelling the interactions between any number of entities. By specifying the input feature vector, \cite{libFM} shows that FM can subsume many specific factorization models such as MF, parallel factor analysis, and SVD++~\cite{SVD++}. As such, FM is recognized as the most effective linear embedding method for sparse data prediction. Many variants to FM have been proposed, such as the neural FM~\cite{He_NFM} that deepens FM under the neural framework to learn high-order feature interactions, and the field-aware FM~\cite{juan2016field} that associates multiple embedding vectors for a feature to differentiate its interaction with other features of different fields.

In this work, we contribute improvements of FM by discriminating the importance of feature interactions. 
We are aware of a work similar to our proposal --- GBFM~\cite{cheng2014gradient}, which selects ``good'' features with gradient boosting and models only the interactions between good features. 
For interactions between selected features, GBFM sums them up with the same weight as FM does. As such, GBFM is essentially a feature selection algorithm, which is fundamentally different with our AFM that can learn the importance of each feature interaction. 

Along another line, deep neural networks (\aka deep learning) are becoming increasingly popular and have recently been employed to prediction under sparse settings. 
Specifically, \cite{cheng2016wide} proposes Wide\&Deep for App recommendation, where the Deep component is a MLP on the concatenation of feature embedding vectors to learn feature interactions; and \cite{shan2016deep} proposes DeepCross for click-through rate prediction, which applies a deep residual MLP~\cite{cvpr16best} to learn cross features. 
We point out that in these methods, feature interactions are implicitly captured by a deep neural network, rather than FM that explicitly models each interaction as the inner product of two features. 
As such, these deep methods are not interpretable, as the contribution of each feature interaction is unknown. 
By directly extending FM with the attention mechanism that learns the importance of each feature interaction, our AMF is more interpretable and empirically demonstrates superior performance over Wide\&Deep and DeepCross.


\section{Experiments}
\label{sec:experiments}
We conduct experiments to answer the following questions: 
\begin{itemize}
	\item[\textbf{RQ1}] How do the key hyper-parameters of AFM (\ie dropout on feature interactions and regularization on the attention network) impact its performance?\vspace{-5pt}
	\item[\textbf{RQ2}] Can the attention network effectively learn the importance of feature interactions?\vspace{-5pt}
	\item[\textbf{RQ3}] How does AFM perform as compared to the state-of-the-art methods for sparse data prediction?\vspace{-5pt}
\end{itemize}

\subsection{Experimental Settings}
\textbf{Datasets.} We perform experiments with two public datasets: Frappe
~\cite{FrappeData} and MovieLens\footnote{\url{grouplens.org/datasets/movielens/latest}}~\cite{MovielensData}. 
The Frappe dataset has been used for context-aware recommendation, which contains $96,203$ app usage logs of users under different contexts. The eight context variables are all categorical, including weather, city, daytime and so on. We convert each log (user ID, app ID and context variables) to a feature vector via one-hot encoding, obtaining $5,382$ features. 
The MovieLens data has been used for personalized tag recommendation, which contains $668,953$ tag applications of users on movies. 
We convert each tag application (user ID, movie ID and tag) to a feature vector and obtain $90,445$ features. 

\noindent \textbf{Evaluation Protocol.} For both datasets, each log is assigned a target of value 1, meaning the user has used the app under the context or applied the tag on the movie. We randomly pair two negative samples with each log and set their target to $-1$.
As such, the final experimental data for Frappe and MovieLens contain $288,609$ and $2,006,859$ instances, respectively. We randomly split each dataset into three portions: 70\% for training, 20\% for validation, and 10\% for testing. The validation set is only used for tuning hyper-parameters, and the performance comparison is done on the test set. To evaluate the performance, we adopt  
\textit{root mean square error} (RMSE), where a lower score indicates a better performance.

\noindent \textbf{Baselines.} We compare AFM with the following competitive methods that are designed for sparse data prediction: 

- \textbf{LibFM}~\cite{libFM}. This is the official C++ implementation for FM. We choose the SGD learner as other methods are all optimized by SGD (or its variants). 

- \textbf{HOFM}. This is the TensorFlow implementation\footnote{\url{https://github.com/geffy/tffm}} of the higher-order FM~\cite{blondel2016higher}. We set the order size to 3, as the MovieLens data has only three types of predictor variables (user, item, and tag). 

- \textbf{Wide\&Deep}~\cite{cheng2016wide}. We implement the method. As the structure (\eg depth and size of each layer) of a deep neural network is difficult to be fully tuned, we use the same structure as reported in the paper. The wide part is the same as the linear regression part of FM, and the deep part is a three-layer MLP with the layer size 1024, 512 and 256.

- \textbf{DeepCross}~\cite{shan2016deep}. We implement the method with the same structure of the original paper. It stacks 5 residual units (each unit has two layers) with hidden dimension 512, 512, 256, 128 and 64, respectively. \\ \vspace{-5pt}

All models are learned by optimizing the squared loss for a fair comparison. 
Besides LibFM, all methods are learned by the mini-batch Adagrad. 
The batch size for Frappe and MovieLens is set to 128 and 4096, respectively. 
The embedding size is set to 256 for all methods. 
Without special mention, the attention factor is also 256, same as the embedding size. 
We carefully tuned the $L_2$ regularization for LibFM and HOFM, and the dropout ratio for Wide\&Deep and DeepCross. 
Early stopping strategy is used based on the performance on validation set.
For Wide\&Deep, DeepCross and AFM, we find that pre-training their feature embeddings with FM leads to a lower RMSE than a random initialization. As such, we report their performance with pre-training. 

\begin{figure}[t]
	\centering
	\begin{minipage}[b]{0.235\textwidth}
		\centering
		\includegraphics[width=\textwidth]{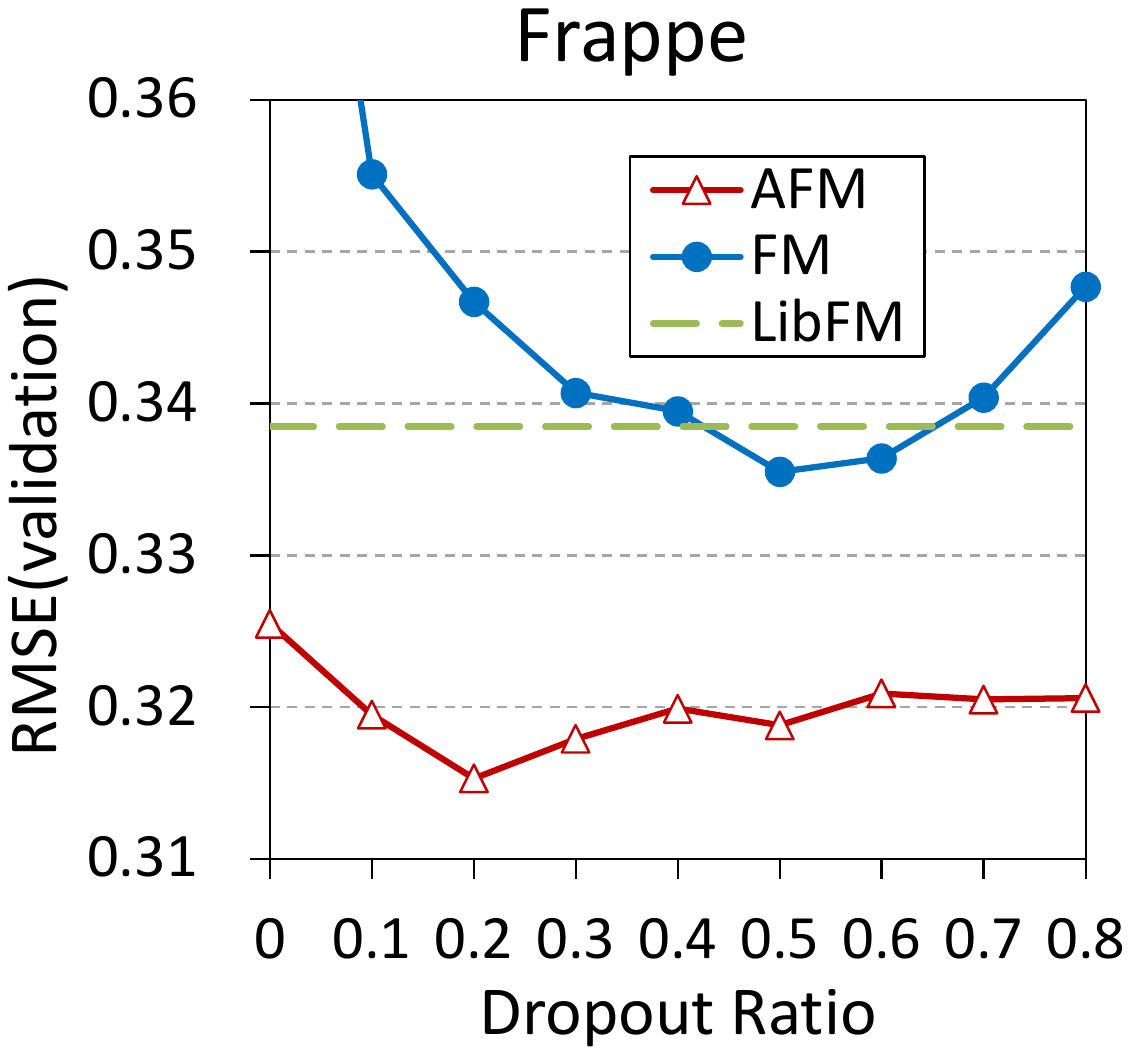}
		\vspace{-18pt}
		\label{fig:dropout_frappe}
	\end{minipage} 
	\begin{minipage}[b]{0.235\textwidth}
		\centering
		\includegraphics[width=\textwidth]{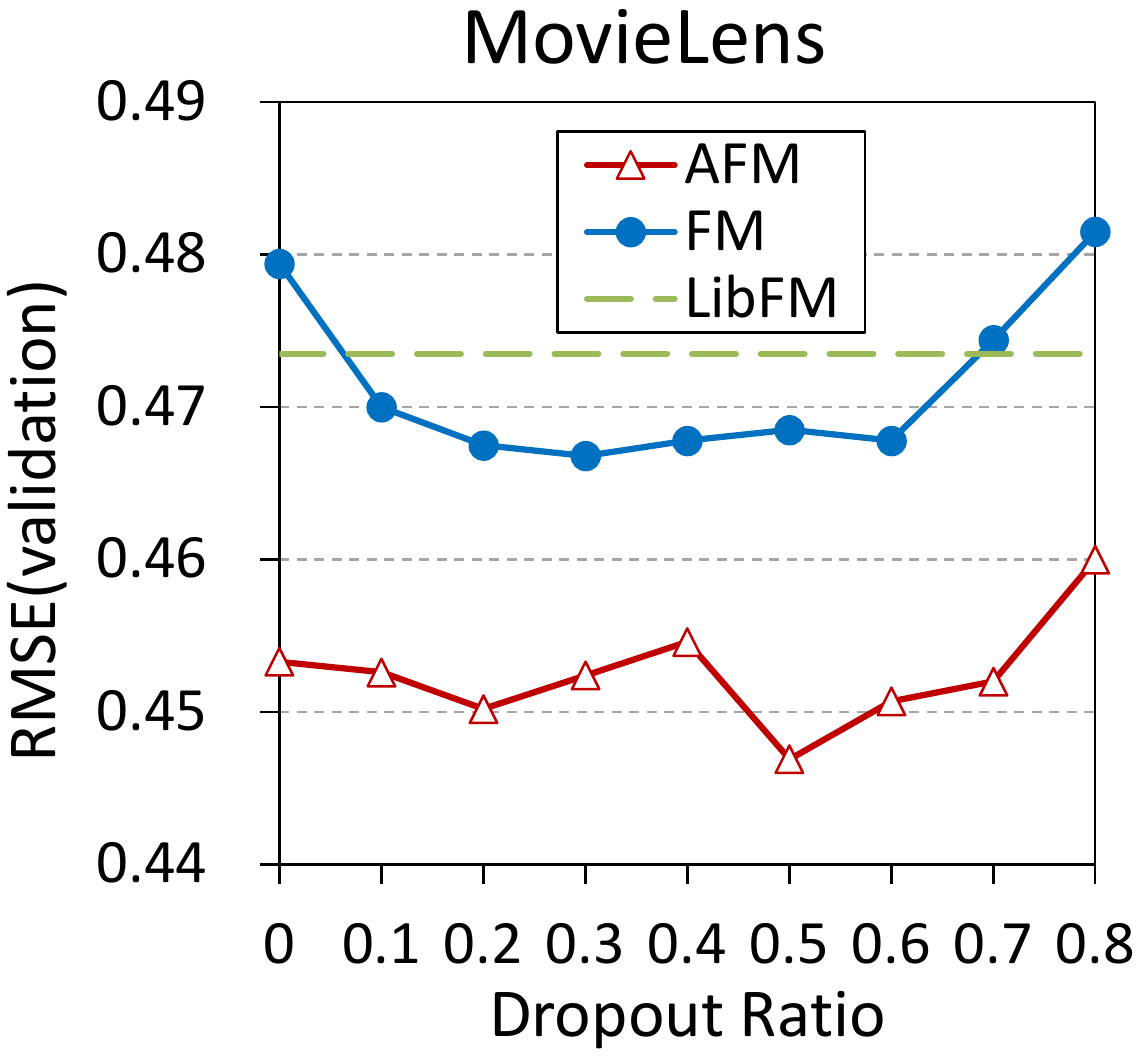}
		\vspace{-18pt}
		\label{fig:dropout_ml_tag}
	\end{minipage} 
	\caption{Validation error of AFM and FM \wrt different dropout ratios on the pair-wise interaction layer}
	\vspace{-15pt}
	\label{fig:dropout}
\end{figure}

\subsection{Hyper-parameter Investigation (RQ1)}
First, we explore the effect of dropout on the pair-wise interaction layer. 
We set $\lambda$ to 0, so that no $L_2$ regularization is used on the attention network. 
We also validate dropout on our implementation of FM by removing the attention component of AFM. 
Figure \ref{fig:dropout} shows the validation error of AFM and FM \wrt different dropout ratios; the result of LibFM is also shown as a benchmark. We have the following observations:
\begin{itemize}[leftmargin=*]
\item By setting the dropout ratio to a proper value, both AFM and FM can be significantly improved. 
Specifically, for AFM, the optimal dropout ratio on Frappe and MovieLens is 0.2 and 0.5, respectively. 
This verifies the usefulness of dropout on the pair-wise interaction layer, which improves the generalization of FM and AFM. 
\item Our implementation of FM offers a better performance than LibFM. The reasons are twofold. First, LibFM optimizes with the vanilla SGD, which adopts a fixed learning rate for all parameters; while we optimize FM with Adagrad, which adapts the learning rate for each parameter based on its frequency (\ie smaller updates for frequent and larger updates for infrequent parameters). Second, LibFM prevents overfitting via $L_2$ regularization, while we employ dropout, which can be more effective due to the model averaging effect. 
\item AFM outperforms FM and LibFM by a large margin. Even when dropout is not used and the overfitting issue does exist to a certain extent, AFM achieves a performance significantly better than the optimal performance of LibFM and FM (\cf the result of dropout ratio equals to 0). This demonstrates the benefits of the attention network in learning the weight of feature interactions. 
\end{itemize}

\begin{figure}[t]
	\centering
	\begin{minipage}[b]{0.235\textwidth}
		\centering
		\includegraphics[width=\textwidth]{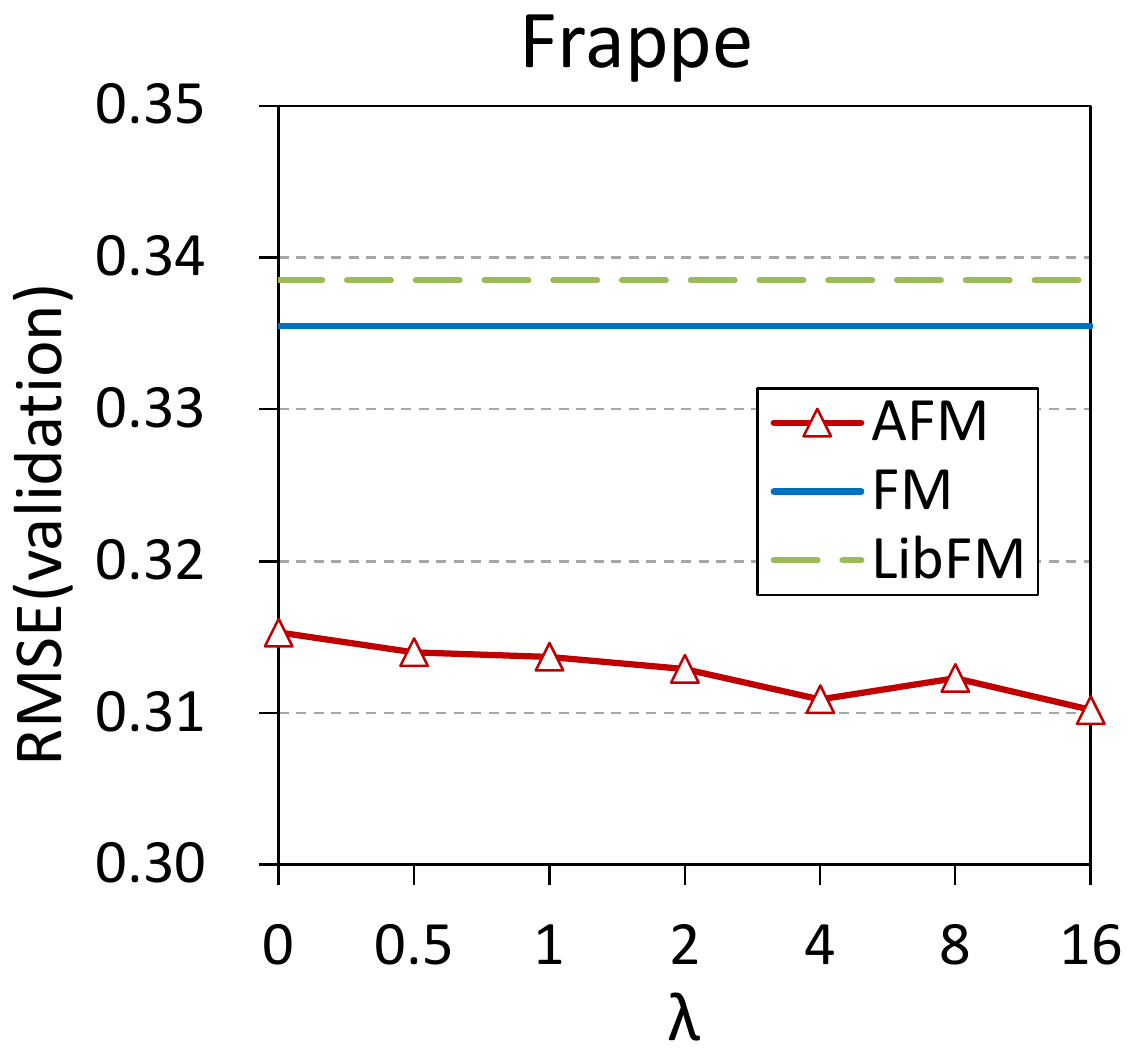}
		\vspace{-18pt}
		\label{fig:lamda_frappe}
	\end{minipage}
	\begin{minipage}[b]{0.235\textwidth}
		\centering
		\includegraphics[width=\textwidth]{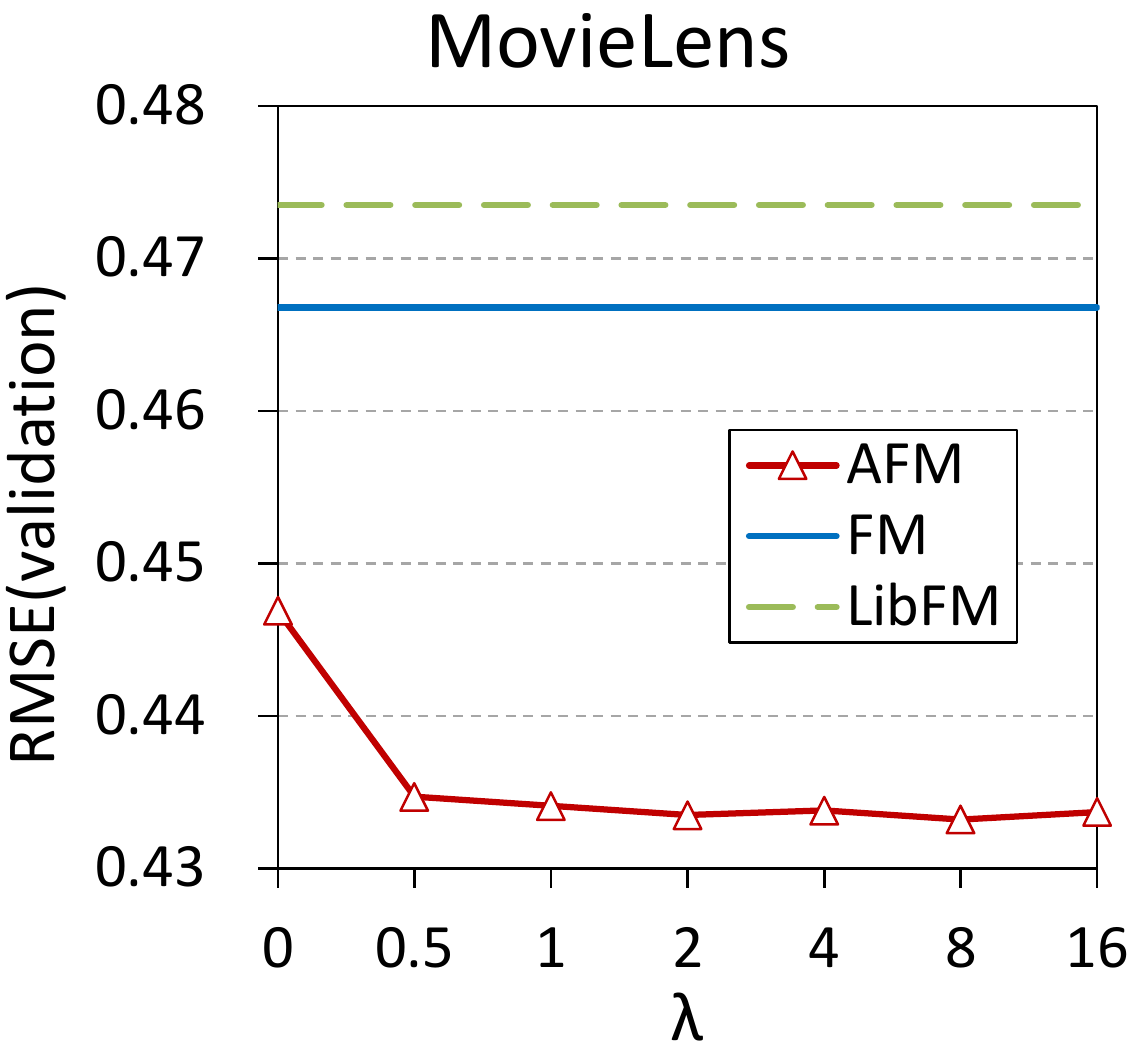}
		\vspace{-18pt}
		\label{fig:lamda_ml_tag}
	\end{minipage}
	\caption{Validation error of AFM \wrt different regularization strengths on the attention network}
	\vspace{-8pt}
	\label{fig:lamda}
\end{figure}

\noindent We then study whether the $L_2$ regularization on the attention network is beneficial to AFM. The dropout ratio is set to the optimal value for each dataset, as evidenced by the previous experiment. As can be seen from Figure~\ref{fig:lamda}, when $\lambda$ is set to a value larger than 0, AFM is improved (note that the result of $\lambda=0$ corresponds to the best performance obtained by AFM in Figure~\ref{fig:dropout}).
This implies that simply using dropout on the pair-wise interaction layer is insufficient to prevent overfitting for AFM. And more importantly, tuning the attention network can further improve the generalization of AFM. 

\subsection{Impact of the Attention Network (RQ2)}
\begin{figure}[t]
	\centering
	\begin{minipage}[b]{0.235\textwidth}
		\centering
		\includegraphics[width=\textwidth]{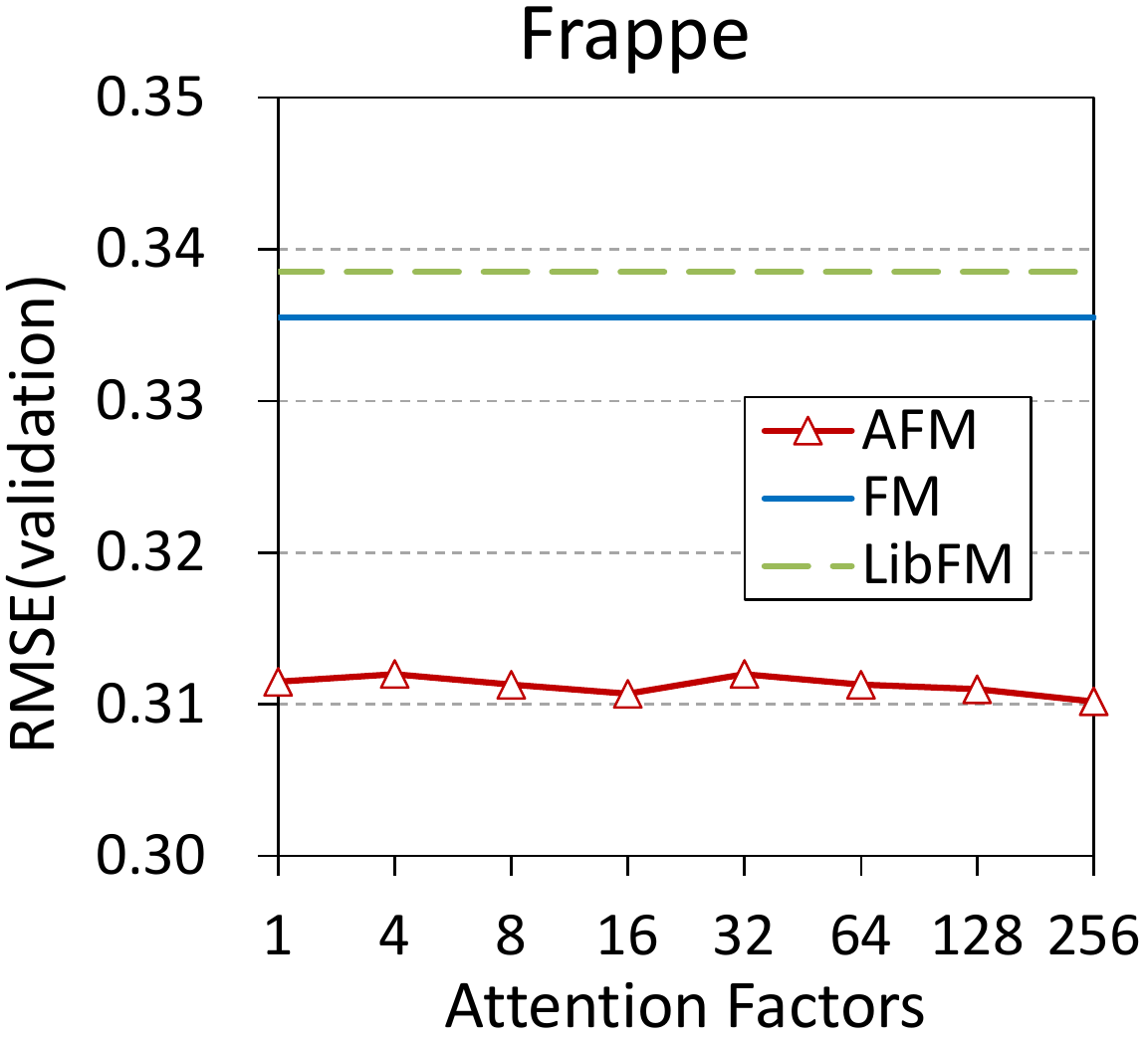}
		\vspace{-18pt}
		\label{fig:attention_factor_frappe}
	\end{minipage} 
	\begin{minipage}[b]{0.235\textwidth}
		\centering
		\includegraphics[width=\textwidth]{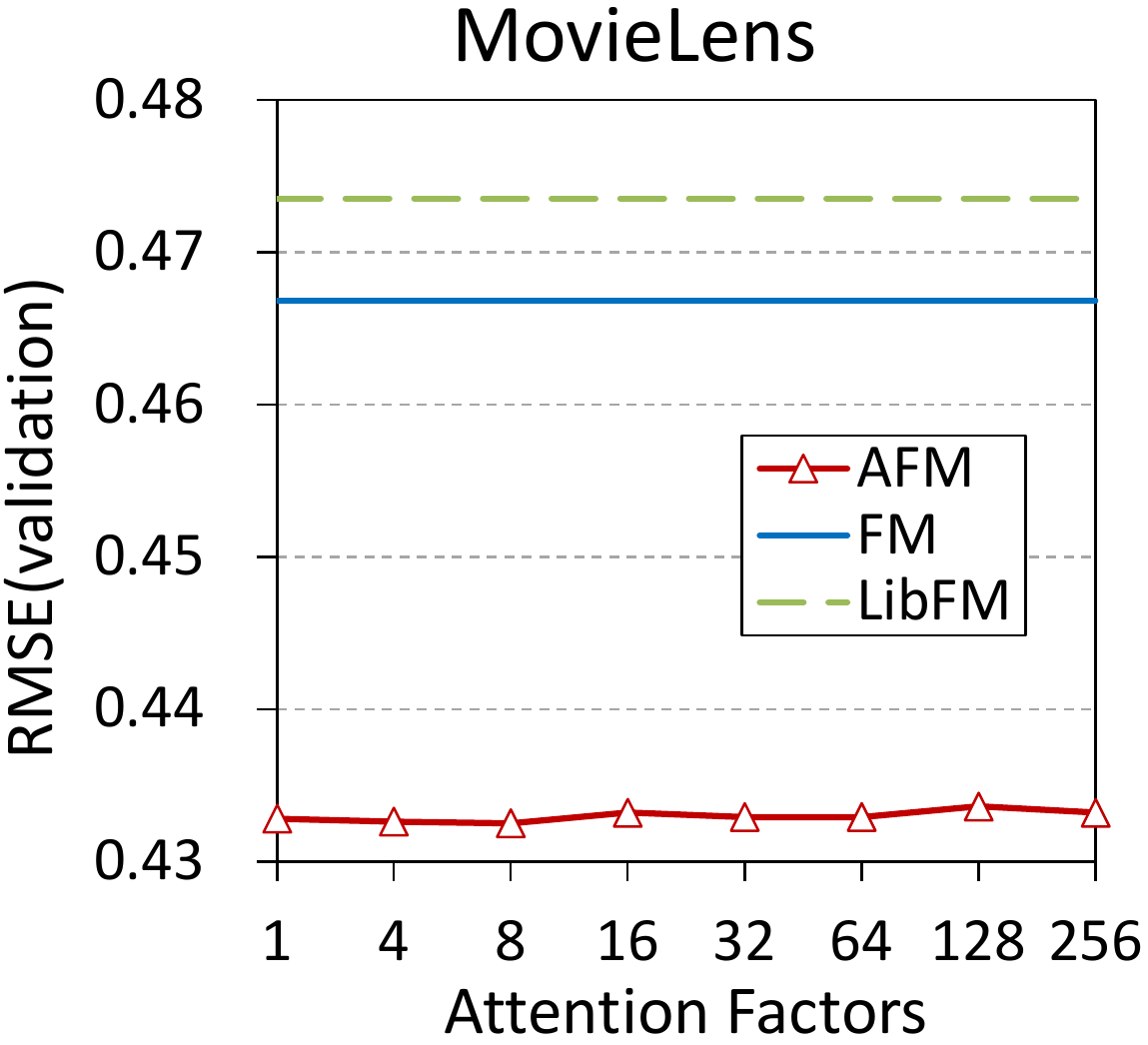}
		\vspace{-18pt}
		\label{fig:attention_factor_ml_tag}
	\end{minipage} 
	\caption{Validation error of AFM \wrt different attention factors}
	\vspace{-15pt}
	\label{fig:attention_factor}
\end{figure}

We now focus on analyzing the impact of the attention network on AFM. The first question to answer is how to choose a proper attention factor? Figure \ref{fig:attention_factor} shows the validation error of AFM \wrt different attention factors. Note that $\lambda$ has been separately tuned for each attention factor. We can observe that for both datasets, AFM's performance is rather stable across attention factors. 
Specifically, when the attention factor is $1$, the $\textbf{W}$ matrix becomes a vector and the attention network essentially degrades to a linear regression model with the interacted vector~(\ie $\textbf{v}_i \odot \textbf{v}_j$) as input features. 
Despite such restricted model capability of the attention component, AFM remains to be very strong and significantly improves over FM. 
This justifies the rationality of AFM's design that estimates the importance score of a feature interaction based on its interacted vector, which is the key finding of this work. 

\begin{figure}[t]
	\centering
	\begin{minipage}[b]{0.235\textwidth}
		\centering
		\includegraphics[width=\textwidth]{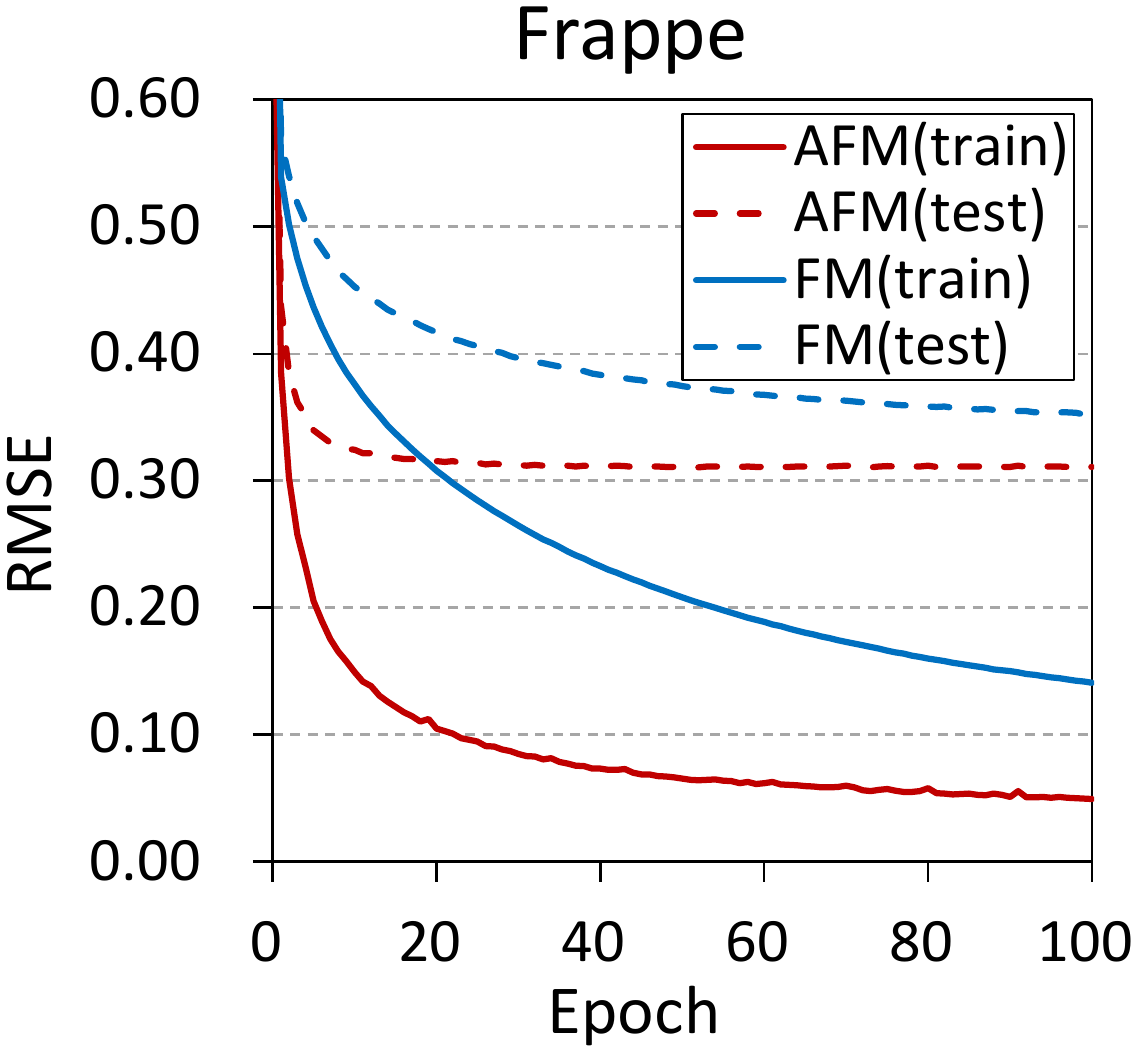}
		\vspace{-18pt}
		\label{fig:epoch_frappe}
	\end{minipage} 
	\begin{minipage}[b]{0.235\textwidth}
		\centering
		\includegraphics[width=\textwidth]{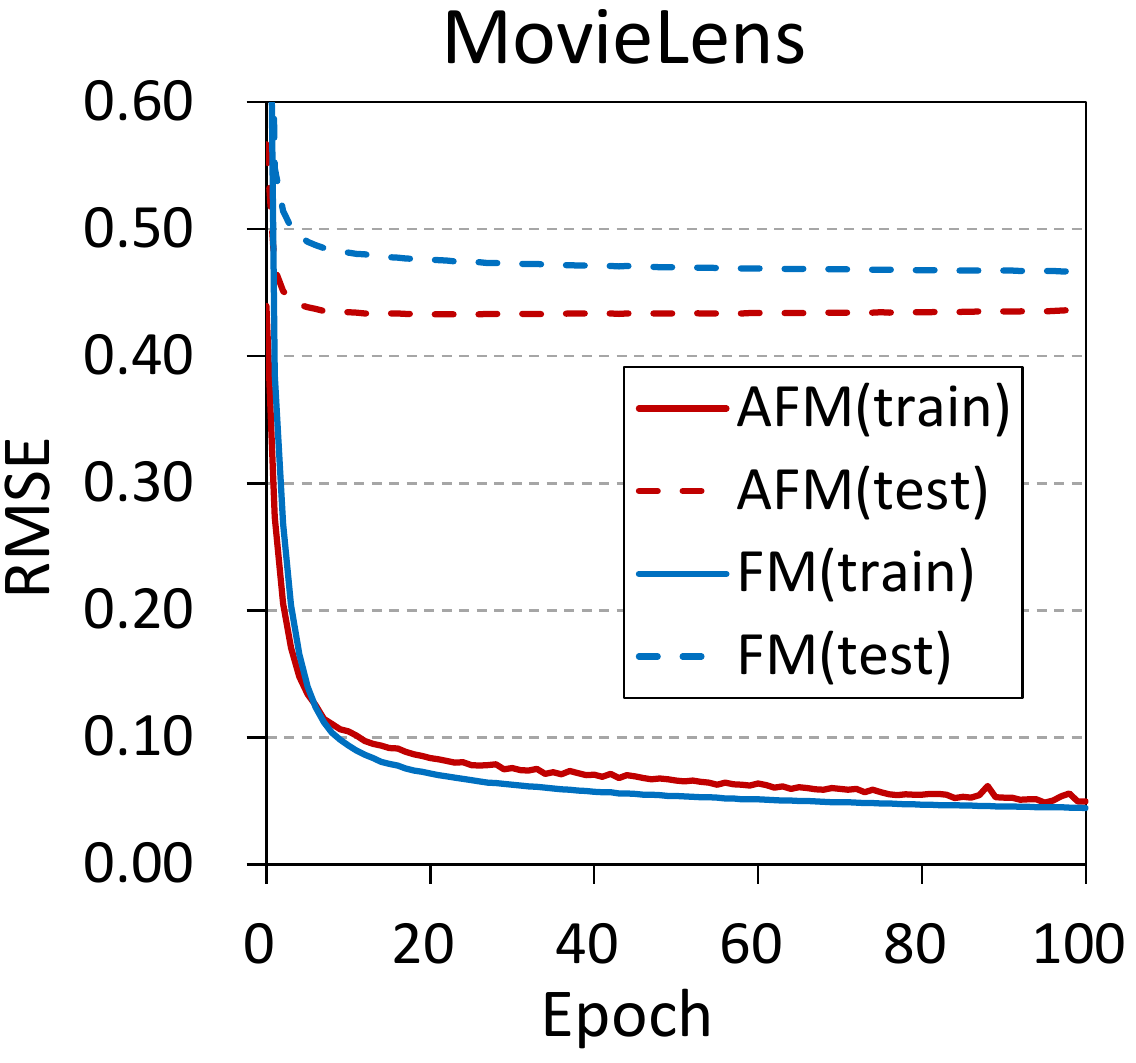}
		\vspace{-18pt}
		\label{fig:epoch_ml_tag}
	\end{minipage} 
	\caption{Training and test error of each epoch}
	\vspace{-15pt}
	\label{fig:epoch}
\end{figure}

Figure \ref{fig:epoch} compares the training and test error of AFM and FM of each epoch. We observe that AFM converges faster than FM. On Frappe, both the training and test error of AFM are much lower than that of FM, indicating that AFM can better fit the data and lead to more accurate prediction. 
On MovieLens, although AFM achieves a slightly higher training error than FM, the lower test error shows that AFM generalizes better to unseen data. 

\subsubsection{Micro-level Analysis}
Besides the improved performance, another key advantage of AFM is that it is more explainable through interpreting the attention score of each feature interaction. To demonstrate this, we perform some micro-level analysis by investigating the score of each feature interaction on MovieLens. 

To allow a dedicated analysis on the attention network, 
we first fix $a_{ij}$ to a uniform number $1/|\mathcal{R}_x|$, training the model which simulates FM. We then fix feature embeddings, training the attention network only; upon convergence, the performance is improved about $3\%$, which justifies the efficacy of the attention network. 
We then select three test examples of target value 1, showing the attention score and interaction score of each feature interaction in Table \ref{tab:case}. We can see that among all three interactions, the item--tag interaction is the most important. 
However, FM assigns the same importance score for all interactions, resulting in a large prediction error. 
By augmenting FM with the attention network (\cf rows FM+A), the item-tag interaction is assigned a higher importance score, and the prediction error is reduced. \vspace{-5pt}

\begin{table}[h]
	\small
	\begin{center}
		\caption{The attention\_score*interaction\_score of each feature interaction of three test examples on MovieLens.}
		\vspace{-10pt}
		\label{tab:case}
		\begin{tabular}{| l | l | c | c | c | c |} \hline
			\# & \textbf{Model} & \textbf{User-Item} & \textbf{User-Tag} & \textbf{Item-Tag} & $\hat{y}$ \\ \hline\hline  
			\multirow{2}{*}{1}  & FM & 0.33*-1.81 & 0.33*-2.65 & 0.33*4.55 & 0.03 \\ \cline{2-6}
			& FM+A & 0.34*-1.81 & 0.27*-2.65 & 0.38*4.55 & 0.39 \\ \hline\hline
			\multirow{2}{*}{2}  & FM & 0.33*-1.62 & 0.33*-1.00 & 0.33*3.32 & 0.24 \\ \cline{2-6}
			& FM+A & 0.38*-1.62 & 0.20*-1.00 & 0.42*3.32 & 0.56 \\ \hline\hline
			\multirow{2}{*}{3}  & FM & 0.33*-1.40 & 0.33*-1.26 & 0.33*4.68 & 0.67 \\ \cline{2-6}
			& FM+A & 0.33*-1.40 & 0.29*-1.26 & 0.37*4.68 & 0.89 \\ \hline
		\end{tabular}
	\end{center}
	\vspace{-15pt}
\end{table}

\subsection{Performance Comparison (RQ3)}
In this final subsection, we compare the performance of different methods on the test set. Table~\ref{tab:performance} summarizes the best performance obtained on embedding size 256 and the number of trainable parameters of each method. 
\begin{itemize}[leftmargin=*]
\item First, we see that AFM achieves the best performance among all methods. Specifically, AFM betters LibFM with a $8.6\%$ relative improvement by using less than $0.1$M additional parameters; and AFM outperforms the second best method Wide\&Deep with $4.3\%$, while using much fewer model parameters. This demonstrates the effectiveness of AFM, which, despite being a shallow model, achieves better performance than deep learning methods.
\item Second, HOFM improves over FM, which is attributed to its modelling of higher-order feature interactions. However, the slight improvements are based on the rather expensive cost of almost doubling the number of parameters, as HOFM uses a separated set of embeddings to model the feature interactions of each order. 
This points to a promising direction of future research --- devising more effective methods for capturing higher-order feature interactions. 
\item Lastly, DeepCross performs the worst, due to the severe problem of overfitting. We find that dropout does not work well for DeepCross, which might be caused by its use of batch normalization. Considering that DeepCross is the deepest method (that stacks 10 layers above the embedding layer) among all compared methods, it provides evidence that deeper leaning is not always helpful, as deep networks can suffer from overfitting and are more difficult to optimize in practice~\cite{He_NFM}.
\end{itemize}

\begin{table}[t]
	\small
	\begin{center}
		\caption{Test error and number of parameters of different methods on embedding size 256. M denotes ``million''.}
		\vspace{-10pt}
		\label{tab:performance}
		\begin{tabular}{| l | c | c | c | c |} \hline
			& \multicolumn{2}{c|}{\textbf{Frappe}} & \multicolumn{2}{c|}{\textbf{MovieLens}} \\ \hline 
			\textbf{Method} & \textbf{Param\#} & \textbf{RMSE} & \textbf{Param\#} & \textbf{RMSE} \\ \hline\hline
			LibFM  & 1.38M  & 0.3385 & 23.24M  & 0.4735 \\ \hline
			HOFM & 2.76M & 0.3331 & 46.40M & 0.4636 \\ \hline\hline
			Wide\&Deep &4.66M & 0.3246  & 24.69M  & 0.4512 \\ \hline
			DeepCross & 8.93M & 0.3548  & 25.42M  & 0.5130  \\ \hline \hline
			\textbf{AFM} & \textbf{1.45M}  & $\textbf{0.3102}$ & \textbf{23.26M}  & $\textbf{0.4325}$\\ \hline
		\end{tabular}
	\end{center}
	\vspace{-10pt}
\end{table}
\section{Conclusion and Future Work}
\label{sec:conclusion}

We have presented a simple yet effective model AFM for supervised learning. Our AFM enhances FM by learning the importance of feature interactions with an attention network, which not only improves the representation ability but also the interpretability of a FM model. This work is orthogonal with our recent work on neural FM~\cite{He_NFM} that develops deep variants of FM for modelling high-order feature interactions, and it is the time that introduces the attention mechanism to factorization machines. 

In future, we will explore deep version for AFM by stacking multiple non-linear layers above the attention-based pooling layer and see whether it can further improve the performance.
As AFM has a relatively high complexity quadratic to the number of non-zero features, we will consider improving its learning efficiency, for example by using learning to hash~\cite{DCF:2016,shen2015supervised} and data sampling~\cite{meng_big_graph} techniques.
Another promising direction is to develop FM variants for semi-supervised and multi-view learning, for example by incorporating the widely used graph Laplacian~\cite{BiRank,meng_semi_supervised} and co-regularization designs~\cite{CoNMF,yang_multitask}. 
Lastly, we will explore AFM on modelling other types of data for different applications, such as texts for question answering~\cite{zhao2015expertQA} and more semantic-rich multi-media content~\cite{zhang2016learning,yang_semantic_videos}. 


\noindent \textbf{Acknowledgment}
The work is supported by the National Natural Science
Foundation of China under Grant No.U1611461 and
No.61572431, Key Research and Development Plan of Zhejiang
Province under Grant No.2015C01027, Zhejiang Natural
Science Foundation under Grant No.LZ17F020001.
NExT research is supported by the National Research Foundation, 
Prime Minister's office, Singapore under its IRC@SG Funding Initiative.

\bibliographystyle{named}

\begin{thebibliography}{}

\end{thebibliography}


\begin{thebibliography}{}
		
		\bibitem[\protect\citeauthoryear{Baltrunas \bgroup \em et al.\egroup
		}{2015}]{FrappeData}
		Linas Baltrunas, Karen Church, Alexandros Karatzoglou, and Nuria Oliver.
		\newblock Frappe: Understanding the usage and perception of mobile app
		recommendations in-the-wild.
		\newblock {\em CoRR}, abs/1505.03014, 2015.
		
		\bibitem[\protect\citeauthoryear{Bayer \bgroup \em et al.\egroup }{2017}]{iCD}
		Immanuel Bayer, Xiangnan He, Bhargav Kanagal, and Steffen Rendle.
		\newblock A generic coordinate descent framework for learning from implicit
		feedback.
		\newblock In {\em WWW}, 2017.
		
		\bibitem[\protect\citeauthoryear{Blondel \bgroup \em et al.\egroup
		}{2016}]{blondel2016higher}
		Mathieu Blondel, Akinori Fujino, Naonori Ueda, and Masakazu Ishihata.
		\newblock Higher-order factorization machines.
		\newblock In {\em NIPS}, 2016.
		
		\bibitem[\protect\citeauthoryear{Chen \bgroup \em et al.\egroup
		}{2016}]{Chen:2016}
		Tao Chen, Xiangnan He, and Min-Yen Kan.
		\newblock Context-aware image tweet modelling and recommendation.
		\newblock In {\em MM}, 2016.
		
		\bibitem[\protect\citeauthoryear{Chen \bgroup \em et al.\egroup
		}{2017a}]{Attentive_CF}
		Jingyuan Chen, Hanwang Zhang, Xiangnan He, Liqiang Nie, Wei Liu, and Tat-Seng
		Chua.
		\newblock Attentive collaborative filtering: Multimedia recommendation with
		feature- and item-level attention.
		\newblock In {\em SIGIR}, 2017.
		
		\bibitem[\protect\citeauthoryear{Chen \bgroup \em et al.\egroup
		}{2017b}]{chen2017cvpr}
		Long Chen, Hanwang Zhang, Jun Xiao, Liqiang Nie, Jian Shao, and Tat{-}Seng
		Chua.
		\newblock {SCA-CNN:} spatial and channel-wise attention in convolutional
		networks for image captioning.
		\newblock In {\em CVPR}, 2017.
		
		\bibitem[\protect\citeauthoryear{Cheng \bgroup \em et al.\egroup
		}{2014}]{cheng2014gradient}
		Chen Cheng, Fen Xia, Tong Zhang, Irwin King, and Michael~R Lyu.
		\newblock Gradient boosting factorization machines.
		\newblock In {\em RecSys}, 2014.
		
		\bibitem[\protect\citeauthoryear{Cheng \bgroup \em et al.\egroup
		}{2016}]{cheng2016wide}
		Heng-Tze Cheng, Levent Koc, Jeremiah Harmsen, et~al.
		\newblock Wide \& deep learning for recommender systems.
		\newblock In {\em DLRS}, 2016.
		
		\bibitem[\protect\citeauthoryear{Harper and Konstan}{2015}]{MovielensData}
		F.~Maxwell Harper and Joseph~A. Konstan.
		\newblock The movielens datasets: History and context.
		\newblock {\em ACM TIIS}, 2015.
		
		\bibitem[\protect\citeauthoryear{He \bgroup \em et al.\egroup
		}{2016a}]{cvpr16best}
		Kaiming He, Xiangyu Zhang, Shaoqing Ren, and Jian Sun.
		\newblock Deep residual learning for image recognition.
		\newblock In {\em CVPR}, 2016.
		
		\bibitem[\protect\citeauthoryear{He \bgroup \em et al.\egroup }{2016b}]{fastMF}
		Xiangnan He, Hanwang Zhang, Min-Yen Kan, and Tat-Seng Chua.
		\newblock Fast matrix factorization for online recommendation with implicit
		feedback.
		\newblock In {\em SIGIR}, 2016.
		
		\bibitem[\protect\citeauthoryear{He \bgroup \em et al.\egroup }{2017a}]{BiRank}
		Xiangnan He, Ming Gao, Min-Yen Kan, and Dingxian Wang.
		\newblock {BiRank}: Towards ranking on bipartite graphs.
		\newblock {\em IEEE TKDE}, 2017.
		
		\bibitem[\protect\citeauthoryear{He \bgroup \em et al.\egroup
		}{2017b}]{He:WWW2017}
		Xiangnan He, Lizi Liao, Hanwang Zhang, Liqiang Nie, Xia Hu, and Tat-Seng Chua.
		\newblock Neural collaborative filering.
		\newblock In {\em WWW}, 2017.
		
		\bibitem[\protect\citeauthoryear{Juan \bgroup \em et al.\egroup
		}{2016}]{juan2016field}
		Yuchin Juan, Yong Zhuang, Wei-Sheng Chin, and Chih-Jen Lin.
		\newblock Field-aware factorization machines for ctr prediction.
		\newblock In {\em RecSys}, 2016.
		
		\bibitem[\protect\citeauthoryear{Koren}{2008}]{SVD++}
		Yehuda Koren.
		\newblock Factorization meets the neighborhood: A multifaceted collaborative
		filtering model.
		\newblock In {\em KDD}, 2008.
		
		\bibitem[\protect\citeauthoryear{Petroni \bgroup \em et al.\egroup
		}{2015}]{petroni2015core}
		Fabio Petroni, Luciano Del~Corro, and Rainer Gemulla.
		\newblock Core: Context-aware open relation extraction with factorization
		machines.
		\newblock In {\em EMNLP}, 2015.
		
		\bibitem[\protect\citeauthoryear{Rendle \bgroup \em et al.\egroup
		}{2011}]{fastFM}
		Steffen Rendle, Zeno Gantner, Christoph Freudenthaler, and Lars Schmidt-Thieme.
		\newblock Fast context-aware recommendations with factorization machines.
		\newblock In {\em SIGIR}, 2011.
		
		\bibitem[\protect\citeauthoryear{Rendle}{2010}]{FM}
		Steffen Rendle.
		\newblock Factorization machines.
		\newblock In {\em ICDM}, 2010.
		
		\bibitem[\protect\citeauthoryear{Rendle}{2012}]{libFM}
		Steffen Rendle.
		\newblock Factorization machines with libfm.
		\newblock {\em ACM TIST}, 2012.
				
		\bibitem[\protect\citeauthoryear{He and Chua}{2017}]{He_NFM}
		Xiangnan He and Tat-Seng Chua.
		\newblock Neural factorization machines for sparse predictive analytics.
		\newblock In {\em SIGIR}, 2017.
		
		\bibitem[\protect\citeauthoryear{He \bgroup \em et al.\egroup }{2014}]{CoNMF}
		Xiangnan He, Min-Yen Kan, Peichu Xie, and Xiao Chen.
		\newblock Comment-based multi-view clustering of web 2.0 items.
		\newblock In {\em WWW}, 2014.
		
		\bibitem[\protect\citeauthoryear{Shan \bgroup \em et al.\egroup
		}{2016}]{shan2016deep}
		Ying Shan, T~Ryan Hoens, Jian Jiao, Haijing Wang, Dong Yu, and JC~Mao.
		\newblock Deep crossing: Web-scale modeling without manually crafted
		combinatorial features.
		\newblock In {\em KDD}, 2016.
		
		\bibitem[\protect\citeauthoryear{Shen \bgroup \em et al.\egroup
		}{2015}]{shen2015supervised}
		Fumin Shen, Chunhua Shen, Wei Liu, and Heng Tao~Shen.
		\newblock Supervised discrete hashing.
		\newblock In {\em CVPR}, 2015.
		
		\bibitem[\protect\citeauthoryear{Srivastava \bgroup \em et al.\egroup
		}{2014}]{srivastava2014dropout}
		Nitish Srivastava, Geoffrey~E Hinton, Alex Krizhevsky, Ilya Sutskever, and
		Ruslan Salakhutdinov.
		\newblock Dropout: a simple way to prevent neural networks from overfitting.
		\newblock {\em JMLR}, 2014.
		
		\bibitem[\protect\citeauthoryear{Wang \bgroup \em et al.\egroup
		}{2015}]{wang2015visual}
		Meng Wang, Xueliang Liu, and Xindong Wu.
		\newblock Visual classification by l1-hypergraph modeling.
		\newblock {\em IEEE TKDE}, 2015.
		
		\bibitem[\protect\citeauthoryear{Wang \bgroup \em et al.\egroup
		}{2016}]{meng_semi_supervised}
		Meng Wang, Weijie Fu, Shijie Hao, Dacheng Tao, and Xindong Wu.
		\newblock Scalable semi-supervised learning by efficient anchor graph
		regularization.
		\newblock {\em IEEE TKDE}, 2016.
		
		\bibitem[\protect\citeauthoryear{Wang \bgroup \em et al.\egroup
		}{2017a}]{silkroad}
		Xiang Wang, Xiangnan He, Liqiang Nie and Tat-Seng Chua 
		\newblock Item Silk Road: Recommending Items from Information Domains to Social Users 
		\newblock {\em SIGIR}, 2017.
		
		\bibitem[\protect\citeauthoryear{Wang \bgroup \em et al.\egroup
		}{2017b}]{meng_big_graph}
		Meng Wang, Weijie Fu, Shijie Hao, Hengchang Liu, and Xindong Wu.
		\newblock Learning on big graph: Label inference and regularization with anchor
		hierarchy.
		\newblock {\em IEEE TKDE}, 2017.
		
		\bibitem[\protect\citeauthoryear{Xiong \bgroup \em et al.\egroup
		}{2017}]{xiong_attend}
		Chenyan Xiong, Jimie Callan, and Tie-Yen Liu.
		\newblock Learning to attend and to rank with word-entity duets.
		\newblock In {\em SIGIR}, 2017.
		
		\bibitem[\protect\citeauthoryear{Yang \bgroup \em et al.\egroup
		}{2014}]{yang_semantic_videos}
		Yang Yang, Zheng-Jun Zha, Yue Gao, Xiaofeng Zhu, and Tat-Seng Chua.
		\newblock Exploiting web images for semantic video indexing via robust
		sample-specific loss.
		\newblock {\em IEEE TMM}, 2014.
		
		\bibitem[\protect\citeauthoryear{Yang \bgroup \em et al.\egroup
		}{2015}]{yang_multitask}
		Yang Yang, Zhigang Ma, Yi~Yang, Feiping Nie, and Heng~Tao Shen.
		\newblock Multitask spectral clustering by exploring intertask correlation.
		\newblock {\em IEEE TCYB}, 2015.
		
		\bibitem[\protect\citeauthoryear{Zhang \bgroup \em et al.\egroup
		}{2016a}]{zhang2016learning}
		Hanwang Zhang, Xindi Shang, Huanbo Luan, Meng Wang, and Tat-Seng Chua.
		\newblock Learning from collective intelligence: Feature learning using social
		images and tags.
		\newblock {\em TMM}, 2016.
		
		\bibitem[\protect\citeauthoryear{Zhang \bgroup \em et al.\egroup
		}{2016b}]{DCF:2016}
		Hanwang Zhang, Fumin Shen, Wei Liu, Xiangnan He, Huanbo Luan, and Tat-Seng
		Chua.
		\newblock Discrete collaborative filtering.
		\newblock In {\em SIGIR}, 2016.
		
		\bibitem[\protect\citeauthoryear{Zhang \bgroup \em et al.\egroup
		}{2017}]{zhang2017relation}
		Hanwang Zhang, Zawlin Kyaw, Shih-Fu Chang, and Tat-Seng Chua.
		\newblock Visual translation embedding network for visual relation detection.
		\newblock In {\em CVPR}, 2017.
		
		\bibitem[\protect\citeauthoryear{Zhao \bgroup \em et al.\egroup
		}{2015}]{zhao2015expertQA}
		Zhou Zhao, Lijun Zhang, Xiaofei He, and Wilfred Ng.
		\newblock Expert finding for question answering via graph regularized matrix
		completion.
		\newblock {\em TKDE}, 2015.
		
		\bibitem[\protect\citeauthoryear{Zhao \bgroup \em et al.\egroup
		}{2016}]{zhao2016user}
		Zhou Zhao, Hanqing Lu, Deng Cai, Xiaofei He, and Yueting Zhuang.
		\newblock User Preference Learning for Online Social Recommendation.
		\newblock {\em TKDE}, 2016.
		
	\end{thebibliography}
\small{

	}

\end{document}